\pdfoutput=1

\documentclass[11pt]{article}

\usepackage[]{acl}

\usepackage{times}
\usepackage{latexsym}
\usepackage{booktabs}
\usepackage{multirow}
\usepackage{tabularx}
\usepackage{xspace}
\usepackage{xcolor,colortbl}
\usepackage{soul}
\usepackage{enumitem}

\usepackage{pifont}
\newcommand{\cmark}{\ding{51}}%
\newcommand{\xmark}{\ding{55}}%
\usepackage[utf8]{inputenc}
\usepackage{CJKutf8}
\newcommand{\zh}[1]{\begin{CJK}{UTF8}{gbsn}#1\end{CJK}}

\usepackage{microtype}
\usepackage{amsmath}
\usepackage{amssymb}
\usepackage{bm}
\usepackage{todonotes}
\usepackage{graphicx}
\usepackage{comment}
\makeatletter
\def\thickhline{%
  \noalign{\ifnum0=`}\fi\hrule \@height \thickarrayrulewidth \futurelet
   \reserved@a\@xthickhline}
\def\@xthickhline{\ifx\reserved@a\thickhline
               \vskip\doublerulesep
               \vskip-\thickarrayrulewidth
             \fi
      \ifnum0=`{\fi}}
\makeatother
\newlength{\thickarrayrulewidth}
\newcommand{\mz}{\textsc{Multi$^{2}$WOZ}\xspace}
\newcommand{\mzs}{\mz\,}

\usepackage{todonotes}
\makeatletter
\newcommand*\iftodonotes{\if@todonotes@disabled\expandafter\@secondoftwo\else\expandafter\@firstoftwo\fi}
\makeatother

\title{Multi$^{2}$WOZ: A Robust Multilingual Dataset and Conversational Pretraining for Task-Oriented Dialog}

\author{Chia-Chien Hung\textsuperscript{1}, Anne Lauscher\textsuperscript{2}, Ivan Vuli\'{c}\textsuperscript{3}, \\ \textbf{Simone Paolo Ponzetto\textsuperscript{1}} and \textbf{Goran Glava\v{s}\textsuperscript{4}} \\
  \textsuperscript{1}Data and Web Science Group, University of Mannheim, Germany \\
  \textsuperscript{2}MilaNLP, Bocconi University, Italy \\
  \textsuperscript{3} LTL, University of Cambridge, UK \hspace{1em} 
  \textsuperscript{4} CAIDAS, University of Würzburg, Germany \\ 
  \texttt{\{chia-chien, simone\}@informatik.uni-mannheim.de} \\ \texttt{anne.lauscher@unibocconi.it}
  \hspace{2em}\texttt{iv250@cam.ac.uk} \\  \texttt{goran.glavas@uni-wuerzburg.de}}
\begin{document}
\maketitle
\begin{abstract}
Research on (multi-domain) task-oriented dialog (TOD) has predominantly focused on the \textit{English} language, primarily due to the shortage of robust TOD datasets in other languages, preventing the systematic investigation of cross-lingual transfer for this crucial NLP application area.     
%
In this work, we introduce \mz, a new multilingual multi-domain TOD dataset, derived from the well-established English dataset \textsc{MultiWOZ}, that spans four typologically diverse languages: Chinese, German, Arabic, and Russian. In contrast to concurrent efforts \cite{ding-etal-2021-globalwoz,zuo-etal-2021-allwoz}, \mzs contains gold-standard dialogs in target languages that are directly comparable with development and test portions of the English dataset, enabling reliable and comparative estimates of cross-lingual transfer performance for TOD.  
We then introduce a new framework for \textit{multilingual conversational specialization} of pretrained language models (PrLMs) that aims to facilitate cross-lingual transfer for arbitrary downstream TOD tasks. Using such conversational PrLMs specialized for concrete target languages, we systematically benchmark a number of zero-shot and few-shot cross-lingual transfer approaches on two standard TOD tasks: Dialog State Tracking and Response Retrieval. Our experiments show that, in most setups, the best performance entails the combination of (i) conversational specialization in the target language and (ii) few-shot transfer for the concrete TOD task. Most importantly, we show that our conversational specialization in the target language allows for an exceptionally \textit{sample-efficient few-shot transfer} for downstream TOD tasks.  


\end{abstract}

\section{Introduction}
Task-oriented dialog (TOD) is arguably one of the most popular natural language processing (NLP) application areas~\citep[\textit{inter alia}]{yan2017building, henderson-etal-2019-training}, with more importance recently given to more realistic, and thus, multi-domain conversations \cite{budzianowski-etal-2018-multiwoz,ramadan-etal-2018-large}, in which users may handle more than one task during the conversation, e.g., booking a \textit{taxi} and making a reservation at a \textit{restaurant}. 
Unlike many other NLP tasks \cite[e.g., ][\textit{inter alia}]{hu2020xtreme,liang2020xglue,ponti2020xcopa}, the progress towards \textit{multilingual multi-domain} TOD has been hindered by the lack of sufficiently large and high-quality datasets in languages other than English \cite{budzianowski-etal-2018-multiwoz,zang2020multiwoz} and more recently, Chinese \cite{zhu-etal-2020-crosswoz}. This lack can be attributed to the fact that  creating TOD datasets for new languages from scratch or via translation of English datasets is significantly more expensive and time-consuming than for most other NLP tasks. However, the absence of multilingual datasets that are comparable (i.e., aligned) across languages prevents a reliable estimate of effectiveness of cross-lingual transfer techniques in multi-domain TOD \cite{razumovskaia2021crossing}.         


In order to address these research gaps, in this work we introduce \mz, a reliable and large multilingual evaluation benchmark for multi-domain task-oriented dialog, derived by translating the monolingual English-only MultiWOZ data~\citep{budzianowski-etal-2018-multiwoz, eric-etal-2020-multiwoz} to four linguistically diverse major world languages, each with a different script: Arabic (AR), Chinese (ZH), German (DE), and Russian (RU). 

Compared to the products of concurrent efforts that derive multilingual datasets from English MultiWOZ \cite{ding-etal-2021-globalwoz,zuo-etal-2021-allwoz}, our \mzs is: (1) much \textit{larger} -- we translate all dialogs from development and test portions of the English MultiWOZ (in total 2,000 dialogs containing the total of 29.5K utterances); (2) much more \textit{reliable} -- complete dialogs, i.e., utterances as well as slot-values, have been manually translated (without resorting to error-prone heuristics), and the quality of translations has been validated through quality control steps; and (3) \textit{parallel} -- the same set of dialogs has been translated to all target languages, enabling the direct comparison of the performance of multilingual models and cross-lingual transfer approaches across languages.          


We then use \mzs to benchmark a range of state-of-the-art zero-shot and few-shot methods for cross-lingual transfer in two standard TOD tasks: Dialog State Tracking (DST) and Response Retrieval (RR). As the second main contribution of our work, we propose a general framework for improving performance and sample-efficiency of cross-lingual transfer for TOD tasks. We first leverage the parallel conversational OpenSubtitles corpus \cite{lison-tiedemann-2016-opensubtitles2016} to carry out a conversational specialization of a PrLM for a given target language, irrespective of the downstream TOD task of interest. We then show that this intermediate conversational specialization in the target language (i) consistently improves the DST and RR performance in both zero-shot and few-shot transfer, and (ii) drastically improves sample-efficiency of few-shot transfer.

\section{Multi$^{2}$WOZ}
In this section we describe the construction of the \mzs dataset, providing also details on inter-translator reliability. We then discuss two concurrent efforts in creating multilingual TOD datasets from MultiWOZ and their properties, and emphasize the aspects that make our \mzs a more reliable and useful benchmark for evaluating cross-lingual transfer for TOD. 

%
%
%


\subsection{Dataset Creation}
\label{sec:data}
%


\label{ss:annotation-process}
\paragraph{Language Selection.} We translate all 2,000 dialogs from the development and test portions of the English MultiWOZ 2.1~\citep{eric-etal-2020-multiwoz} dataset to Arabic (AR), Chinese (ZH), German (DE), and Russian (RU). We selected the target languages based on the following criteria: (1) linguistic diversity (DE and RU belong to different Indo-European subfamilies -- Germanic and Slavic, respectively; ZH is a Sino-Tibetan language and AR Semitic), (2) diversity of scripts (DE and RU use Latin and Cyrillic scripts, respectively, both \textit{alphabet} scripts; AR script represents the \textit{Abjad} script type, whereas the ZH Hanzi script belongs to \textit{logographic} scripts), (3) number of native speakers (all four are in the top 20 most-spoken world languages), and (4) our access to native and fluent speakers of those languages who are proficient in English.  

\paragraph{Two-Step Translation.} 
Following the well-established practice, we carried out a two-phase translation of the English data: (1) we started with an \textit{automatic translation} of the dialogs -- utterances as well as the annotated slot values -- followed by (2) the \textit{manual post-editing} of the translations. We first automatically translated all utterances and slot values from the development and test dialogs from the MultiWOZ 2.1~\citep{eric-etal-2020-multiwoz} (1,000 dialogs in each portion; 14,748 and 14,744 utterances, respectively) to our four target languages, using Google Translate.\footnote{Relying on its Python API: \url{https://pypi.org/project/googletrans}}
We then hired two native speakers of each target language,\footnote{In order to reduce the translation costs, we initially attempted to post-edit the translations via crowdsourcing. We tried this for Russian using the popular platform Toloka (\url{toloka.yandex.com}); however, the translation quality remained unsatisfactory even after several post-editing rounds.} all with a University degree and fluent in English, to post-edit the (non-overlapping sets of) automatic translations, i.e., fix the errors in automatic translations of utterances as well as slot values. 

\setlength{\tabcolsep}{3pt}
\begin{table*}[t]
    \centering
    \small{
    \begin{tabular}{l l l}
    \toprule
        & \textbf{Utterance} & \textbf{Value for ``attraction-name''} \\
        \midrule
         \multirow{2}{*}{Original} &  \textit{No hold off on booking for now.}  & \multirow{2}{*}{\textit{cineworld cinema}} \\
         & \textit{Can you help me find an attraction called cineworld cinema?} & \\
         Automatic Trans. & \zh{目前暂无预订。您能帮我找到一个名为\emph{cineworld Cinema}的景点吗？} & \emph{Cineworld}\zh{电影} \\
         Manual Correc. & \zh{目前暂无预订。您能帮我找到一个名为电影世界电影院的景点吗？} &  \zh{电影世界电影院}\\
    \bottomrule
    \end{tabular}}
    \caption{Example utterance (from the dialog MUL0484) with a value for a slot (\textit{``attraction-name''}). We show the original English text, the automatic translation to Chinese and the final translation after manual post-editing.}
    \label{tab:example}
\end{table*}
Since we carried out the automatic translation of the utterances independently of the automatic translation of the slot values, the translators were instructed to pay special attention to the alignment between each translated utterance and translations of slot value annotations for that utterance. We show an example utterance with associated slot values after the automatic translation and manual post-editing in Table~\ref{tab:example}.



\paragraph{Quality Control.} 
In order to reduce the translation costs, our human post-editors worked on disjoint sets of dialogs. Because of this, our annotation process contained an additional quality assurance step. Two new annotators for each target language judged the correctness of the translations on the random sample of 200 dialogs (10\% of all translated dialogs, 100 from the development and test portion each), containing 2,962 utterances in total. The annotators had to independently answer the following questions for each translated utterance from the sample: 
(1) \textit{Is the utterance translation acceptable?} and (2) \textit{Do the translated slot values match the translated utterance?} On average, across all target languages, both quality annotators for the respective language answered affirmatively to both questions for 99\% of all utterances. Adjusting for chance agreement, we measured the Inter-Annotator Agreement (IAA) in terms of Cohen's $\kappa$~\citep{cohen1960coefficient}, observing the almost perfect agreement\footnote{According to \newcite{landis1977measurement}, if $\kappa\geq 0.81$.} of $\kappa=0.824$ for the development set and $\kappa=0.838$ for test set.


\paragraph{Annotation Duration and Cost.} In total, we hired 16 annotators, four for each of our four target languages: two for post-editing and two for quality assessment. The overall effort spanned almost full 5 months (from July to November 2021), and amounted to 1,083 person-hours. With the remuneration rate of 16 \$/h, creating \mzs cost us \$17,328.   


\subsection{Comparison with Concurrent Work} 
\label{ss:absolute_sh_t}
Two concurrent works also derive multilingual datasets from MultiWOZ \cite{ding-etal-2021-globalwoz,zuo-etal-2021-allwoz}, with different strategies and properties, discussed in what follows.

GlobalWOZ \cite{ding-etal-2021-globalwoz} encompasses Chinese, Indonesian, and Spanish datasets. The authors first create \textit{templates} from dialog utterances by replacing slot-value strings in the utterances with the slot type and value index (e.g., \textit{``\dots and the post code is \underline{cb238el}''} becomes the template \textit{``\dots and the post code is \texttt{[attraction-postcode-1]}''}. They then \textit{automatically} translate all templates to the target languages. Next, they select a subset of 500 test set dialogs for human post-editing with the following heuristic: dialogs for which the sum of corpus-level frequencies of their constitutive 4-grams (normalized with the dialog length) is the largest.\footnote{Interestingly, the authors do not provide any motivation or intuition for this heuristic. It is also worth noting that they count the 4-gram frequencies, upon which the selection of the dialogs for post-editing depends, on the noisy automatic translations.} Since this selection step is independent for each language, each GlobalWOZ portion contains translations of a different subset of English dialogs: this prevents any direct comparison of downstream TOD performance across languages. Even more problematically,  the selection heuristic directly reduces linguistic diversity of dialogs chosen for the test set of each language, as it favors the dialogs that contain the same globally most frequent 4-grams. Due to this artificial homogeneity of its test sets, GlobalWOZ is very likely to overestimate downstream TOD performance for target languages. 

Unlike GlobalWOZ, AllWOZ \cite{zuo-etal-2021-allwoz} does automatic translation of a \textit{fixed} \textit{small} subset of MultiWOZ plus post-editing in seven target languages. However, it encompasses only 100 dialogs and 1,476 turns; as such, it is arguably too small to draw strong conclusions about the performance of cross-lingual transfer methods. Its usefulness in joint domain and language transfer evaluations is especially doubtful, since it covers individual MultiWOZ domains with an extremely small number of dialogs (e.g., only 13 for the \textit{Taxi} domain). Finally, neither \newcite{ding-etal-2021-globalwoz} nor \newcite{zuo-etal-2021-allwoz} provide any estimates of the quality of their final datasets nor do they report their annotation costs.        


In contrast to GlobalWOZ, \mzs is a parallel corpus -- with the exact same set of dialogs translated to all four target languages; as such it directly enables performance comparisons across the target languages. Further, containing translations of \textit{all} dev and test dialogs from MultiWOZ (i.e., avoiding sampling heuristics), \mzs does not introduce any confounding factors that would distort estimates of cross-lingual transfer performance in downstream TOD tasks. Finally, \mzs is 20 times larger (per language) than AllWOZ: experiments on \mzs are thus much more likely to yield conclusive findings. 

\section{Cross-lingual Transfer for TOD}
\label{sec:method}
The parallel nature and sufficient size of \mzs allow us to benchmark and compare a number of established and novel cross-lingual transfer methods for TOD. In particular, (1) we first inject general conversational TOD knowledge into XLM-RoBERTa~\citep[XLM-R;][]{conneau-etal-2020-unsupervised}, yielding TOD-XLMR (\S\ref{ss:tod-xlmr}); (2)~we then propose several variants for conversational specialization of TOD-XLMR for target languages, better suited for transfer in downstream TOD tasks (\S\ref{ss:intermediate-training}); (3) we investigate zero-shot and few-shot transfer for two TOD tasks: DST and RR (\S\ref{ss:downstream_transfer}).


\subsection{TOD-XLMR: A Multilingual TOD Model}
\label{ss:tod-xlmr}
Recently, \newcite{wu-etal-2020-tod} demonstrated that specializing BERT~\citep{devlin-etal-2019-bert} on conversational data by means of additional pretraining via a combination of masked language modeling (MLM) and response selection (RS) objectives yields improvements in downstream TOD tasks. 
Following these findings, we first (propose to) conversationally specialize XLM-R \citep{conneau-etal-2020-unsupervised}, a state-of-the-art multilingual PrLM covering 100 languages, in the same manner: applying the RS and MLM objectives on the same English conversational corpus consisting of nine human-human multi-turn TOD datasets (see~\citet{wu-etal-2020-tod} for more details). As a result, we obtain TOD-XLMR -- a massively multilingual PrLM specialized for task-oriented conversations. Note that TOD-XLMR is not yet specialized (i.e., fine-tuned) for any concrete TOD task (e.g., DST or Response Generation). Rather, it is enriched with general task-oriented conversational knowledge (in English), presumed to be beneficial for a wide variety of TOD tasks. 

\vspace{-0.5em}
\begin{table*}[!t]
\def\arraystretch{0.93}
\resizebox{\textwidth}{!}
{
\begin{tabular}{lllllll}
\toprule
\multicolumn{3}{c|}{\textbf{EN Subtitle}} & \multicolumn{3}{c}{\textbf{ZH Subtitle}} \\
\midrule
\multicolumn{3}{l|}{\begin{tabular}[c]{@{}l@{}}- Professor Hall. - Yes. - I think your theory may be correct. - Walk with me. \\ Just a few weeks ago, I monitored the strongest hurricane on record. \\ The hail, the tornados, it all fits. \\ Can your model factor in storm scenarios?\end{tabular}} & \multicolumn{3}{l}{\begin{tabular}[c]{@{}l@{}}\zh{-霍尔教授 -是的 -我认为你的理论正确 -跟我来} \\ \zh{上周我观测到史上最大的飓风} \\ \zh{雹暴和龙卷风也符合你的理论} \\ \zh{你能预测暴风雨的形成吗？}\end{tabular}} \\
\midrule
\midrule
\textbf{Translation LM (TLM)} & \multicolumn{5}{|l}{- Professor Hall. - Yes. - I think your theory may be [MASK]. - Walk with...\zh{-霍尔教授 -是的 -我认为你的[MASK]正确...}} \\
\midrule
\textbf{Response Selection (RS)} & \multicolumn{2}{|l}{\begin{tabular}[c]{@{}l@{}} \textit{Context:}\\ \zh{上周我观测到史上最大的飓风}\\ \zh{雹暴和龙卷风也符合你的理论}\end{tabular}} 
& \multicolumn{2}{l}{\begin{tabular}[c]{@{}l@{}}{\textbf{Monolingual (RS-Mono)}} \\\textit{True Response:}\\ \zh{你能预测暴风雨的形成吗？}\\ \textit{False Response:}\\ \zh{你有彼得的电脑断层扫描吗？}\end{tabular}}
& \multicolumn{2}{l}{\begin{tabular}[c]{@{}l@{}}{\textbf{Cross-lingual (RS-X)}} \\ \textit{True Response:}\\ Can your model factor in storm scenarios?\\ \textit{False Response:}\\ Do you have Peter's CT scan results?\end{tabular}}\\
\bottomrule
\end{tabular}%
}
\caption{Examples of training instances for conversational specialization for the target language created from OpenSubtitles (OS). Top row: an example of a dialog created from OS, parallel in English and Chinese. Below are training examples for different training objectives: (1) \textit{Translation Language Modelling} (\textbf{TLM}) on the interleaved English-Chinese parallel utterances; (2) two variants of \textit{Response Selection} (RS) -- (a) monolingual in the target language (\textbf{RS-Mono}) and (b) cross-lingual (\textbf{RS-X}).
}
\label{tab:os-example}
\vspace{-0.8em}
\end{table*}

\subsection{Target-Language Specialization}
\label{ss:intermediate-training}
TOD-XLMR has been conversationally specialized only in English data. We next hypothesize that a further conversational specialization for a concrete target language X can improve the transfer EN$\rightarrow$X for all downstream TOD tasks. Accordingly, similar to \citet{moghe-etal-2021-cross}, we investigate several intermediate training procedures that further conversationally specialize TOD-XLMR for the target language X (or jointly for EN and X).       
%
For this purpose, we (i) compile target-language-specific as well as cross-lingual corpora from the CCNet \cite{wenzek-etal-2020-ccnet} and OpenSubtitles \cite{lison-tiedemann-2016-opensubtitles2016} datasets and (ii) experiment with different monolingual, bilingual, and cross-lingual training procedures. Here, we propose a novel cross-lingual response selection (RS) objective and demonstrate its effectiveness in cross-lingual transfer for downstream TOD tasks. 

\paragraph{Training Corpora.}
\label{ss:data-collection}
We collect two types of data for language specialization: (i) \textit{``flat'' corpora} (i.e., without any conversational structure): we simply randomly sample 100K sentences for each language from the respective monolingual portion of CCNet (we denote with \textit{Mono-CC} the individual 100K-sentence portions of each language; with \textit{Bi-CC} the concatenation of the English and each of target language\textit{ Mono-CCs}, and with \textit{Multi-CC} the concatenation of all five Mono-CC portions); (ii) \textit{parallel dialogs} (in EN and target language X) from OpenSubtitles (OS), a parallel conversational corpus spanning 60 languages, compiled from subtitles of movies and TV series. We leverage the parallel OS dialogs to create two different cross-lingual specialization objectives, as described next.

\paragraph{Training Objectives.} 
\label{ss:objectives}
We directly use the CC portions (Mono-CC, Bi-CC, and Multi-CC) for standard \textbf{MLM} training. We then leverage the parallel OS dialogs for two training objectives. First, we carry out translation language modeling (\textbf{TLM}) \cite{conneau2019cross} on the synthetic dialogs which we obtain by interleaving $K$ randomly selected English utterances with their respective target language translations; we then (as with MLM), dynamically mask 15\% of tokens of such interleaved dialogs; we vary the size of the context the model can see when predicting missing tokens by randomly selecting $K$ (between $2$ and $15$) for each instance. Second, we use OS to create instances for both monolingual and cross-lingual Response Selection (RS) training. RS is a simple binary classification task in which for a given pair of a \textit{context} (one or more consecutive utterances) and \textit{response} (a single utterance), the model has to predict whether the response utterance immediately follows the context (i.e., it is a \emph{true} response) or not (i.e., it is a \textit{false} response). RS pretraining has been proven beneficial for downstream TOD in monolingual English setups \cite{mehri-etal-2019-pretraining,henderson-etal-2019-training,henderson-etal-2020-convert,hung-etal-2021-dstod}. 

In this work, we leverage the parallel OS data to introduce the cross-lingual RS objective, where the context and the response utterance are not in the same language. In our experiments, we carry out both (i) monolingual RS training in the target language (i.e., both the context and response utterance are, e.g., in Chinese), denoted \textbf{RS-Mono}, and (ii) cross-lingual RS between English (as the source language in downstream TOD tasks) and the target language, denoted \textbf{RS-X}. We create \textit{hard RS negatives}, by coupling contexts with non-immediate responses from the same movie or episode (same \texttt{imdbID}), as well as \textit{easy negatives} by randomly sampling $m \in \{1, 2, 3\}$ responses from a different movie of series episode (i.e., different \texttt{imdbID}). Hard negatives encourage the model to reason beyond simple lexical cues. Examples of training instances for OS-based training (for EN-ZH) are shown in Table~\ref{tab:os-example}.

\subsection{Downstream Cross-lingual Transfer}
\label{ss:downstream_transfer}
Finally, we fine-tune the various variants of TOD-XLMR, obtained through the above-described specialization (i.e., intermediate training) procedures, for two downstream TOD tasks (DST and RR) and examine their cross-lingual transfer performance. We cover two cross-lingual transfer scenarios: (1) \textit{zero-shot transfer} in which we only fine-tune the models on the English training portion of MultiWOZ and evaluate their performance on the \mzs test data of our four target languages; and (2) \textit{few-shot transfer} in which we sequentially first fine-tune the models on the English training data and then on the small number of dialogs from the development set of \mz, in similar vein to \cite{lauscher-etal-2020-zero}. In order to determine the effect of our conversational target language specialization (\S\ref{ss:intermediate-training}) on the downstream sample efficiency, we run few-shot experiments with different numbers of target language training dialogs, ranging from 1\% to 100\% of the size of \mzs development portions.       
\section{Experimental Setup}
\label{s:setup}

\paragraph{Evaluation Tasks and Measures.} 
We evaluate different multilingual conversational PrLMs in cross-lingual transfer (zero-shot and few-shot) for two prominent TOD tasks: \emph{dialog state tracking (DST)} and \emph{response retrieval (RR)}. 

DST is commonly cast as a multi-class classification task, where given a predefined ontology and dialog history (a sequence of utterances), the model has to predict the output state, i.e., \textit{(domain, slot, value)} tuples~\citep{wu-etal-2020-tod}.\footnote{The model is required to predict slot values for each \textit{(domain, slot)} pair at each dialog turn.} We adopt the standard \textit{joint goal accuracy} as the evaluation measure: at each dialog turn, it compares the predicted dialog states against the manually annotated ground truth which contains slot values for all the \textit{(domain, slot)} candidate pairs. A prediction is considered correct if and only if all predicted slot values exactly match the ground truth. 

RR is a ranking task that is well-aligned with the RS objective and relevant for retrieval-based TOD systems~\citep{wu-etal-2017-sequential,henderson-etal-2019-training}: given the dialog context, the model ranks $N$ dataset utterances, including the \textit{true response} to the context (i.e., the candidate set includes the one \textit{true} response and $N$-1 \textit{false} responses). We follow \citet{henderson-etal-2020-convert} and report the results for $N = 100$, i.e., the evaluation measure is recall at the top $1$ rank given $99$ randomly sampled false responses, denoted as $\textsc{R}_{100}@1$.

\paragraph{Models and Baselines.} 
We briefly summarize the models that we compare in zero-shot and few-shot cross-lingual transfer for DST and RR. 
As baselines, we report the performance of the vanilla multilingual PrLM XLM-R~\citep{conneau-etal-2020-unsupervised}\footnote{We use \texttt{xlm-roberta-base} from HuggingFace.} and its variant further trained on the English TOD data from \cite{wu-etal-2020-tod}: TOD-XLMR (\S\ref{ss:tod-xlmr}). Comparison between XLM-R and TOD-XLMR quantifies the effect of conversational English pretraining on downstream TOD performance, much like the comparison between BERT and TOD-BERT done by \citet{wu-etal-2020-tod}; however, here we extend the comparison to cross-lingual transfer setups. 
We then compare the baselines against a series of our target language-specialized variants, obtained via intermediate training on CC (Mono-CC, Bi-CC, and Multi-CC) by means of MLM, and on OS jointly via TLM and RS (RS-X or RS-Mono) objectives (see \S\ref{ss:intermediate-training} again).

\paragraph{Hyperparameters and Optimization.} 
For training TOD-XLMR (\S\ref{ss:tod-xlmr}), we 
select the effective batch size of 8. 
In target-language-specific intermediate training (\S\ref{ss:intermediate-training}), we fix the maximum sequence length to $256$ subword tokens; for RS objectives, we limit the context and response to $128$ tokens each. We train for $30$ epochs in batches of size $16$ for MLM/TLM, and $32$ for RS. We search for the optimal learning rate among the following values: $\{10^{-4}, 10^{-5}, 10^{-6}\}$. We apply early stopping based on development set performance (patience:~$3$~epochs for MLM/TLM,~$10$~epochs for RS). 
In downstream fine-tuning, we train in batches of $6$ (DST) and $24$ instances (RR) with the initial learning rate fixed to $5\cdot 10^{-5}$. 
We also apply early stopping (patience: $10$ epochs) based on the development set performance, training maximally for 300 epochs in zero-shot setups, and for 15 epochs in target-language few-shot training. In all experiments, we use Adam~\citep{kingma2014adam} as the optimization algorithm.

\section{Results and Discussion}
\label{sec:results}

We now present and discuss the downstream cross-lingual transfer results on \mz for DST and RR in two different transfer setups: zero-shot transfer and few-shot transfer.
\vspace{-0.5em}
\subsection{Zero-Shot Transfer}

\setlength{\tabcolsep}{4pt}
\label{sec:experiments}
\begin{table}[!t]
\centering
{
\small
\begin{tabularx}{\linewidth}{lccccc}
\toprule
\textbf{Model} &
  \textbf{DE} & \textbf{AR} & \textbf{ZH} & \textbf{RU} & \textbf{Avg.} \\ 
  \midrule
  \multicolumn{6}{l}{\textit{w/o intermediate specialization}} \\ \midrule
  
XLM-R& 
1.41   & 1.15  & 1.35  & 1.40  & 1.33  \\  
TOD-XLMR & 
1.74   & 1.53  & 1.75  & 2.16  & 1.80  \\
\midrule
\multicolumn{6}{l}{\textit{with conversational target-lang.~specialization}} \\ \midrule
MLM on Mono-CC & 3.57 & 2.71  & 3.34  & 5.17  & 3.70  \\ 
\hspace{3.7em} Bi-CC & 3.66 & 2.17  & 2.73  & 3.73  & 3.07 \\ 
\hspace{3.7em} Multi-CC &  3.65   & 2.35  & 2.06  & 5.39  & 3.36 \\ 
TLM on OS & 7.80   & 2.43  & 3.95  & 6.03  & 5.05 \\ 
TLM + RS-X on OS & \textbf{7.84}   & \textbf{3.12}  & 4.14  & 6.13  & 5.31 \\ 
TLM + RS-Mono on OS & 7.67   &  2.85     & \textbf{4.47}  & \textbf{6.57}  & \textbf{5.39} \\ 
\bottomrule
\end{tabularx}
}
\caption{Performance of multilingual conversational models in zero-shot cross-lingual transfer for Dialog State Tracking (DST) on \mz, with joint goal accuracy (\%) as the evaluation metric. Reference English DST performance of TOD-XLMR: 47.86\%.}
\label{tab:eval_result_dst_zs}
\vspace{-0.5em}
\end{table}

\paragraph{Dialog State Tracking.} Table \ref{tab:eval_result_dst_zs} summarizes zero-shot cross-lingual transfer performance for DST. First, we note that the transfer performance of all models for all four target languages is extremely low, drastically lower than the reference English DST performance of TOD-XLMR, which stands at 47.9\%. These massive performance drops, stemming from cross-lingual transfer are in line with findings from concurrent work \cite{ding-etal-2021-globalwoz,zuo-etal-2021-allwoz} and suggest that reliable cross-lingual transfer for DST is much more difficult to achieve than for most other language understanding tasks \cite{hu2020xtreme,ponti2020xcopa}. 

Despite low performance across the board, we do note a few emerging and consistent patterns. First, TOD-XLMR slightly but consistently outperforms the vanilla XLM-R, indicating that \textit{conversational} English pretraining brings marginal gains. All of our proposed models from \S\ref{ss:intermediate-training} (the lower part of Table \ref{tab:eval_result_dst_zs}) substantially outperform TOD-XLMR, proving that intermediate conversational specialization for the target language brings gains, irrespective of the training objective. 

Expectedly, TLM and RS training on parallel OS data brings substantially larger gains than MLM-ing on flat monolingual target-language corpora (Mono-CC) or simple concatenations of corpora from two (Bi-CC) or more languages (Multi-CC). German and Arabic seem to benefit slightly more from the cross-lingual Response Selection training (RS-X), whereas for Chinese and Russian we obtain better results with the monolingual (target language) RS training (RS-Mono).

\setlength{\tabcolsep}{4pt}
\begin{table}[!t]
\centering
{\small
\begin{tabularx}{\linewidth}{lccccc}
\toprule
\textbf{Model} &
  \textbf{DE} & \textbf{AR} & \textbf{ZH} & \textbf{RU} & \textbf{Avg.} \\ 
  \midrule
  \multicolumn{6}{l}{\textit{w/o intermediate specialization}} \\ \midrule
TOD-XLMR & 3.3& 2.9& 1.9& 2.7& 2.7 \\ \midrule
\multicolumn{6}{l}{\textit{with conversational target-lang.~specialization}} \\ \midrule
MLM on Mono-CC & 22.9& 25.5& 24.5& 33.4& 26.6 \\
TLM on OS & \textbf{44.4}& 30.3& 34.1& 39.3& 37.0 \\
TLM + RS-Mono on OS & 44.3 & \textbf{30.9}& \textbf{34.8}& \textbf{39.6}& \textbf{37.4} \\
  \bottomrule
\end{tabularx}%
}
\caption{Performance of multilingual conversational models in zero-shot cross-lingual transfer for Response Retrieval (RR) on \mz with $\textsc{R}_{100}@1$ (\%) as the evaluation metric. Reference English RR performance of TOD-XLMR: 64.75\%}
\label{tab:eval_result_rr}
\vspace{-0.1em}
\end{table}

\paragraph{Response Retrieval.} 
The results of zero-shot transfer for RR are summarized in Table \ref{tab:eval_result_rr}. Compared to DST results, for the sake of brevity, we show the performance of only the stronger baseline (TOD-XLMR) and the best-performing variants with intermediate conversational target-language training (one for each objective type): MLM on Mono-CC, TLM on OS, and TLM\,+\,RS-Mono on OS. Similar to DST, TOD-XLMR exhibits a near-zero cross-lingual transfer performance for RR as well, across all target languages. In sharp contrast to DST results, however, conversational specialization for the target language -- with any of the three specialization objectives -- massively improves the zero-shot cross-lingual transfer for RR. The gains are especially large for the models that employ the parallel OpenSubtitles corpus in intermediate specialization, with the monolingual (target language) Response Selection objective slightly improving over TLM training alone. 

Given the parallel nature of \mz, we can directly compare transfer performance of both DST and RR across the four target languages. In both tasks, the best-performing models exhibit stronger performance (i.e., smaller performance drops compared to the English performance) for German and Russian than for Arabic and Chinese. This aligns well with the linguistic proximity of the target languages to English as the source language.     
\vspace{-0.3em}

\vspace{-0.8em}
\begin{figure*}[t!]
	\centering
    \includegraphics[width=\textwidth]{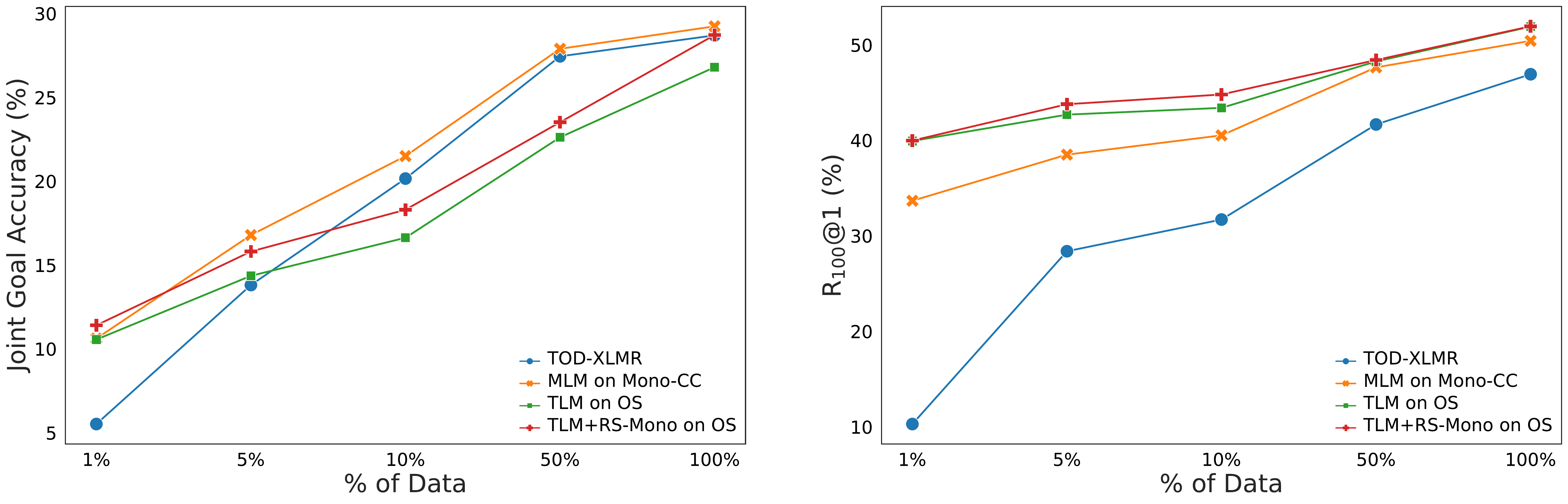}
    \vspace{-0.2em}
	\caption{Few-shot cross-lingual transfer results for Dialog State Tracking (left figure) and Response Retrieval (right figure), averaged across all four target languages (detailed per-language results available in the Appendix). Results shown for different sizes of the training data in the target-language (i.e., different number of \textit{shots}): 1\%, 5\%, 10\%, 50\% and 100\% of the \mzs development sets (of respective target languages).} 
	\label{fig:dst_rr_ave_linept}
\end{figure*}

\setlength{\tabcolsep}{6pt}
\begin{table*}[t]
\centering
{\small
\begin{tabularx}{\linewidth}{clcccccccccc}
\toprule
 &  & \multicolumn{5}{c}{\textbf{DST}} & \multicolumn{5}{c}{\textbf{RR}} \\
 \cmidrule(lr){3-7} \cmidrule(lr){8-12}
\textbf{Lang} & \textbf{Model} & \textbf{1\%} & \textbf{5\%} & \textbf{10\%} & \textbf{50\%} & \textbf{100\%} & \textbf{1\%} & \textbf{5\%} & \textbf{10\%} & \textbf{50\%} & \textbf{100\%} \\
\midrule
\multirow{2}{*}{\textbf{DE}} & TOD-XLMR & 7.68 & 19.26 & 28.08 & 33.17 & 34.10 & 10.25 & 32.47 & 35.56 & 45.39 & 49.46 \\
 & TLM+RS-Mono on OS & 15.88 & 24.14 & 28.38 & 32.57 & 35.45 & 46.08 & 48.94 & 49.98 & 53.43 & 55.72 \\
  \midrule
\multirow{2}{*}{\textbf{AR}} & TOD-XLMR & 1.48 & 1.57 & 6.18 & 15.62 & 17.63 & 6.36 & 18.72 & 23.57 & 36.04 & 42.69 \\
 & TLM+RS-Mono on OS & 4.42 & 6.79 & 8.27 & 14.39 & 21.48 & 33.45 & 37.09 & 38.01 & 41.89 & 47.15 \\
  \midrule
\multirow{2}{*}{\textbf{ZH}} & TOD-XLMR & 8.63 & 12.55 & 16.40 & 23.45 & 25.49 & 15.69 & 31.10 & 33.22 & 41.97 & 48.14 \\
 & TLM+RS-Mono on OS & 11.63 & 14.90 & 17.97 & 22.81 & 28.84 & 38.45 & 43.71 & 45.27 & 48.50 & 51.81 \\
  \midrule
\multirow{2}{*}{\textbf{RU}} & TOD-XLMR & 4.34 & 21.89 & 30.01 & 37.58 & 37.61 & 8.90 & 31.31 & 34.51 & 43.33 & 47.45 \\
 & TLM+RS-Mono on OS & 13.74 & 17.44 & 18.63 & 24.33 & 29.15 & 41.97 & 45.44 & 46.02 & 49.90 & 53.16\\
 \bottomrule
\end{tabularx}%
}
\caption{Per-language few-shot transfer performance (sample efficiency results) on DST and RR for the baseline TOD-XLMR and the best specialized model (TLM+RS-Mono on OS).}
\label{tab:per_lang}
\vspace{-0.5em}
\end{table*}
\subsection{Few-Shot Transfer and Sample Efficiency}
\vspace{-0.3em}
Next, we present the results of few-shot transfer experiments, where we additionally fine-tune the task-specific TOD model on a limited number of target-language dialogs from the development portion of \mz, after first fine-tuning it on the complete English training set from MultiWOZ (see \S\ref{s:setup}). Few-shot cross-lingual transfer results, averaged across all four target languages, are summarized in Figure~\ref{fig:dst_rr_ave_linept}. The figure shows the performance for different sizes of the target-language training data (i.e., number of target-language shots, that is, percentage of the target-language development portion from \mz). Detailed per-language few-shot results are given in Table~\ref{tab:per_lang}, for brevity only for TOD-XLMR and the best target-language-specialized model (TLM+RS-Mono on OS). We provide full per-language results for all specialized models from Figure~\ref{fig:dst_rr_ave_linept} in the Appendix.

The few-shot results unambiguously show that the intermediate conversational specialization for the target language(s) \textit{drastically improves the target-language sample efficiency in the downstream few-shot transfer}. The baseline TOD-XLMR -- not exposed to any type of conversational pretraining for the target language(s) -- exhibits substantially lower performance than all three models (MLM on Mono-CC, TLM on OS, and TLM+RS-Mono on OS) that underwent conversational intermediate training on respective target languages. This is evident even in the few-shot setups where the three models are fine-tuned on merely 1\% (10 dialogs) or 5\% (50 dialogs) of the \mzs development data (after prior fine-tuning on the complete English task data from MultiWOZ). 

As expected, the larger the number of task-specific (DST or RR) training instances in the target languages (50\% and 100\% setups), the closer the performance of the baseline TOD-XLMR gets to the best-performing target-language-specialized model -- this is because the size of the in-language training data for the concrete task (DST or RR) becomes sufficient to compensate for the lack of conversational target-language intermediate training that the specialized models have been exposed to. The sample efficiency of the conversational target-language specialization is more pronounced for RR than for DST. This seems to be in line with the zero-shot transfer results (see Tables \ref{tab:eval_result_dst_zs} and \ref{tab:eval_result_rr}), where the specialized models displayed much larger cross-lingual transfer gains over TOD-XLMR on RR than on DST. We hypothesize that this is due to the intermediate specialization objectives (especially RS) being better aligned with the task-specific training objective of RR than that of DST.                     
\vspace{-0.3em}

\section{Related Work}
\label{sec:rw}
\vspace{-0.3em}
\paragraph{TOD Datasets.}
Research in task-oriented dialog has been, for a long time, limited by the existence of only monolingual English datasets. While earlier datasets focused on a single domain \cite{henderson2014second,henderson2014third,wen-etal-2017-network}, the focus shifted towards the more realistic multi-domain task-oriented dialogs with the creation of the MultiWOZ dataset \cite{budzianowski-etal-2018-multiwoz}, which has been refined and improved in several iterations \cite{eric-etal-2020-multiwoz,zang2020multiwoz,han2021multiwoz}.  
Due to the particularly high costs of creating TOD datasets (in comparison with other language understanding tasks) \cite{razumovskaia2021crossing}, only a handful of monolingual TOD datasets in languages other than English~\citep{zhu-etal-2020-crosswoz} or bilingual TOD datasets have been created~\citep{gunasekara2020overview, lin2021bitod}. \citet{mrksic-etal-2017-semantic} were the first to translate 600 dialogs from the single-domain WOZ 2.0~\citep{mrksic-etal-2017-neural} to Italian and German. 
Concurrent work \citep{ding-etal-2021-globalwoz,zuo-etal-2021-allwoz}, which we discuss in detail in \S\ref{ss:absolute_sh_t} and compare thoroughly against our \mz, introduces the first multilingual multi-domain TOD datasets, created by translating portions of MultiWOZ to several languages. 



\paragraph{Language Specialization and Cross-lingual Transfer.}
Multilingual transformer-based models (e.g., mBERT~\citep{devlin-etal-2019-bert}, XLM-R~\citep{conneau-etal-2020-unsupervised}) are pretrained on large general-purpose and massively multilingual corpora (over 100 languages). While this makes them versatile and widely applicable, it does lead to suboptimal representations for individual languages, a phenomenon commonly referred to as the ``curse of multilinguality''~\citep{conneau-etal-2020-unsupervised}. Therefore, one line of research focused on adapting (i.e., \textit{specializing}) those models to particular languages \cite{lauscher-etal-2020-zero,pfeiffer-etal-2020-mad}. For example, \newcite{pfeiffer-etal-2020-mad} propose a more computationally efficient approach for extending the model capacity for individual languages: this is done by augmenting the multilingual PrLM with language-specific adapter modules.
\citet{glavas-etal-2020-xhate} perform language adaptation through additional intermediate masked language modeling in the target languages with filtered text corpora, demonstrating substantial gains in downstream zero-shot cross-lingual transfer for hate speech and abusive language detection tasks. In a similar vein, \citet{moghe-etal-2021-cross} carry out intermediate fine-tuning of multilingual PrLMs on parallel conversational datasets and demonstrate its effectiveness in zero-shot cross-lingual transfer for the DST task. 

\newcite{lauscher-etal-2020-zero} show that few-shot transfer, in which one additionally fine-tunes the PrLM on a few labeled task-specific target-language instances leads to large improvements for many task-and-language combinations, and that labelling a few target-language examples is more viable than further LM-specialization for languages of interest under strict zero-shot conditions. This finding is also corroborated in our work for two TOD tasks.  
\vspace{-0.3em}
%




\section{Reproducibility}
\label{sec:reproducibility}
\vspace{-0.3em}
To ensure full reproducibility of our results and further fuel research on multilingual TOD, we release the parameters of TOD-XLMR within the Huggingface repository as the first publicly available multilingual PrLM specialized for TOD.\footnote{\url{https://huggingface.co/umanlp/TOD-XLMR}} We also release our code and data and provide the annotation guidelines for \textit{manual post-editing} and \textit{quality control} utilized during the creation of \textsc{Multi$^{2}$WOZ} in the Appendix. This makes our approach completely transparent and fully reproducible.
All resources developed as part of this work are publicly available at: \url{https://github.com/umanlp/Multi2WOZ}. 
\vspace{-0.3em}

\section{Conclusion}
\label{sec:conc}
\vspace{-0.3em}
%

Task-oriented dialog (TOD) has predominantly focused on \textit{English}, primarily due to the lack of robust TOD datasets in other languages \cite{razumovskaia2021crossing}, preventing systematic investigations of cross-lingual transfer methodologies in this crucial NLP application area. To address this gap, in this work, we have presented \textsc{Multi$^{2}$WOZ} -- a robust multilingual multi-domain TOD dataset. \mzs encompasses gold-standard dialogs in four languages (German, Arabic, Chinese, and Russian) that are directly comparable with development and test portions of the English MultiWOZ dataset, thus allowing for the most reliable and comparable estimates of cross-lingual transfer performance for TOD to date. Further, we presented a framework for \textit{multilingual conversational specialization} of pretrained language models that facilitates cross-lingual transfer for downstream TOD tasks. Our experiments on \mzs for two prominent TOD tasks -- Dialog State Tracking and Response Retrieval -- reveal that the cross-lingual transfer performance benefits from both (i) intermediate conversational specialization for the target language and (ii) few-shot cross-lingual transfer for the concrete downstream TOD task. Crucially, we show that our novel conversational specialization for the target language leads to \textit{exceptional sample efficiency} in downstream few-shot transfer. 

In hope to steer and inspire future research on multilingual and cross-lingual TOD, we make \textsc{Multi$^{2}$WOZ} publicly available and will extend the resource to further languages from yet uncovered language families (e.g., Turkish).
\vspace{-0.3em}



\section*{Acknowledgements}
\label{sec:acknow}
\vspace{-0.3em}
The work of Goran Glava\v{s} has been supported by the Multi2ConvAI project of MWK Baden-Württemberg. Simone Paolo Ponzetto has been supported by the JOIN-T 2 project of the Deutsche Forschungsgemeinschaft (DFG). Chia-Chien Hung has been supported by JOIN-T 2 (DFG) and Multi2ConvAI (MWK BW). The work of Anne Lauscher was funded by Multi2ConvAI and by the European Research Council (grant agreement No. 949944, INTEGRATOR). The work of Ivan Vuli\'{c} has been supported by the ERC PoC Grant MultiConvAI (no. 957356) and a Huawei research
donation to the University of Cambridge. 
\vspace{-0.3em}

\section*{Ethical Considerations}
\vspace{-0.3em}
In this work, we have presented \textsc{Multi$^{2}$WOZ}, a robust multilingual multi-domain TOD dataset, and focused on the multilingual conversational specialization of pretrained language models. Although the scope of this manuscript does not allow for an in-depth discussion of the potential ethical issues associated with conversational artificial intelligence in general, we would still like to highlight the ethical sensitivity of this area of NLP research and emphasize some of the potential harms of conversational AI applications, which propagate to our work. For instance, issues may arise from unfair stereotypical biases encoded in general purpose~\cite{lauscher2021sustainable} as well as in conversational~\cite{barikeri-etal-2021-redditbias} pretrained language models and from exclusion of the larger spectrum of (gender) identities~\citep{lauscher2022welcome}. Furthermore, (pre)training as well as fine-tuning of large-scale PrLMs can be hazardous to the environment~\citep{strubell-etal-2019-energy}: in this context, the task-agnostic intermediate conversational specialization for the target languages that we introduce, which allows for highly sample-efficient fine-tuning for various TOD tasks can be seen as a step in the positive direction, towards the reduction of the carbon footprint of neural dialog models. 
\vspace{-0.3em}

\newpage
\bibliography{references}
\bibliographystyle{acl_natbib}

\appendix

\clearpage
\label{sec:appendix}
\section{Annotation Guidelines: Post-editing of the Translation}
\setcounter{section}{0}
\renewcommand{\thesection}{\arabic{section}}

\section{Task Description}
Multi-domain Wizard-of-Oz dataset (MultiWOZ)~\cite{budzianowski-etal-2018-multiwoz} is introduced as a 
fully-labeled collection of human-to-human written conversations spanning over multiple domains and topics. 

Our project aims to translate the monolingual English-only MultiWOZ dataset to four linguistically diverse major world languages, each with a different script: Arabic (AR), Chinese (ZH), German (DE), and Russian (RU). 

In this annotation task, we resort to the revised version 2.1~\cite{eric-etal-2020-multiwoz} and focus on the development and test portions of the English MultiWOZ 2.1 (in total of 2,000 dialogs containing a total of 29.5K utterances). 
We first \textit{automatically translate} all the utterances and the annotated slot values to the four target languages, using Google Translate. Next the translated utterances and slot values (i.e., fix the translation errors) will be \textit{post-edited} with manual efforts. 

For this purpose, a JSON file for \textit{development} or \textit{test} set will be provided to each annotator. There are two tasks: (1) Fix the errors in automatic translations of translated utterances and the translated slot values. (2) Check the alignment between each translated utterance and the slot value annotations for that utterance.

\section{JSON Representation}
The JSON file will be structured as follows, feel free to use any JSON editor tools (e.g., JSON Editor Online) to annotate the files.

\paragraph{Annotation data}
\begin{itemize}
  \item \textbf{dialogID}: An unique ID for each dialog.
  \item \textbf{turnID}: The turn ID of the utterance in the dialog.
  \item \textbf{services}: Domain(s) of the dialog.
  \item \textbf{utterance}: English utterance from MultiWOZ.
  \item \textbf{SlotValues}: English annotated slot values from MultiWOZ.
  \item \textbf{transUtterance}: Translated utterance from Google Translate.
  \item \textbf{transSlotValues}: Translated slot values from Google Translate.
\end{itemize}

\paragraph{Annotation Task}
\begin{itemize}
  \item \textbf{fixTransUtterance}: The revised translated utterance with manual efforts.
  \item \textbf{fixTransSlotValues}: The revised translated slot values with manual efforts.
  \item \textbf{changedUtterance}: Whether the translated utterance is changed. Annotate as 1 if the translated utterance is revised, 0 otherwise.
  \item \textbf{changedSlotValues}: Whether the translated slot values is changed. Annotate as 1 if the translated slot values are revised, 0 otherwise.
\end{itemize}

\section{Annotation Example}

\paragraph{Example 1: Name Correction and Mismatch}\mbox{} \\
The following example in Chinese shows the error fixed with the translated name issue, and also the correctness of the mismatch case between the translated utterance and translated slot values.\vspace{0.7em}
\noindent\fbox{%
    \parbox{0.47\textwidth}{%
    \vspace{0.3em}
        \textbf{dialogID}: \textit{MUL0484.json}\\
        \textbf{turnID}: \textit{6} \\
        \textbf{services}: \textit{train, attraction} \\
        \textbf{utterance}: \textit{No hold off on booking for now. Can you help me find an attraction called cineworld cinema?}\\
        \textbf{slotValues}: {\{attraction-name: \textit{cineworld cinema}\}}\\
        \textbf{transUtterance}: \zh{目前暂无预订。您能帮我找到一个名为cineworld Cinema的景点吗？}\\
        \textbf{transSlotValues}: {\{attraction-name: Cineworld\zh{电影}\}}
        \vspace{0.7em}
        \hrule
        \vspace{0.7em}
        \textbf{fixTransUtterance}: \zh{目前暂无预订。您能帮我找到一个名为电影世界电影院的景点吗？}\\
        \textbf{fixTransSlotValues}: {\{attraction-name: \zh{电影世界电影院}\}}\\
        \textbf{changedUtterance}: 1\\
        \textbf{changedSlotValues}: 1\\
    }%
}

\paragraph{Example 2: Grammatical Error}\mbox{} \\
The following example in German shows the error corrected based on the grammatical issue of the translated utterance.\vspace{0.7em}
\noindent\fbox{%
    \parbox{0.47\textwidth}{%
    \vspace{0.3em}
        \textbf{dialogID}: \textit{PMUL1072.json}\\
        \textbf{turnID}: \textit{6} \\
        \textbf{services}: \textit{train, attraction} \\
        \textbf{utterance}: \textit{I'm leaving from Cambridge.}\\
        \textbf{slotValues}: {\{train-departure: \textit{cambridge}\}}\\
        \textbf{transUtterance}: \textit{Ich verlasse Cambridge.}\\
        \textbf{transSlotValues}: {\{train-departure: \textit{cambridge}\}}
        \vspace{0.7em}
        \hrule
        \vspace{0.7em}
        \textbf{fixTransUtterance}: \textit{Ich fahre von Cambridge aus.}\\
        \textbf{fixTransSlotValues}: {\{train-departure: \textit{cambridge}\}}\\
        \textbf{changedUtterance}: 1\\
        \textbf{changedSlotValues}: 0\\
    }%
}

\section{Additional Notes}
There might be some cases of synonyms. For example, in Chinese \zh{周五} and \zh{星期五} both have the same meaning as \textit{Friday} in English, also similarly in Russian regarding the weekdays. In this case, just pick the most common one and stays consistent among all the translated utterances and slot values. Besides there might be some language variations across different regions, please ignore the dialects and metaphors while fixing the translation errors.

If there are any open questions that you think are not covered in this guide, please do not hesitate to get
in touch with me or post the questions on Slack, so these issues can be discussed together with other annotators and the guide can be improved.

\clearpage
\setcounter{section}{1}
\renewcommand{\thesection}{\Alph{section}}
\section{Annotation Guidelines: Quality Control}
\setcounter{section}{0}
\renewcommand{\thesection}{\arabic{section}}
\section{Task Description}
Multi-domain Wizard-of-Oz dataset (MultiWOZ)~\cite{budzianowski-etal-2018-multiwoz} is introduced as a 
fully-labeled collection of human-to-human written conversations spanning over multiple domains and topics. Our project is aimed to translate the monolingual English-only MultiWOZ dataset to four linguistically diverse major world languages, each with a different script: Arabic (AR), Chinese (ZH), German (DE), and Russian (RU). In the previous annotation task, we resorted to the revised version 2.1~\cite{eric-etal-2020-multiwoz} and focused on the development and test portions of the English MultiWOZ 2.1. 

According to the translation process, it was processed in two steps: we first \textit{automatically translated} all the utterances and the annotated slot values to the four target languages, using Google Translate. Next the translated utterances and slot values (i.e., fix the translation errors) were \textit{post-edited} with manual efforts from native speakers of each language. 

Additionally, a \textit{quality assurance} step is required to check the quality of the post-edited translation. For this purpose, a JSON file for a random sample 200 dialogs (100 from the development and test set each), containing 2,962 utterances in total will be provided to two annotators for each target language to judge the correctness of the translations. Each annotator has to independently answer the following questions for each translated utterance from the sample: (1) \textit{Is the utterance translation acceptable?} (2) \textit{Do the translated slot values match the translated utterance?} 

\paragraph{Annotation data}
\begin{itemize}
  \item \textbf{dialogID}: An unique ID for each dialog.
  \item \textbf{turnID}: The turn ID of the utterance in the dialog.
  \item \textbf{utterance}: English utterance from MultiWOZ.
  \item \textbf{SlotValues}: English annotated slot values from MultiWOZ.
  \item \textbf{fixTransUtterance}: The revised translated utterance with manual efforts.
  \item \textbf{fixTransSlotValues}: The revised translated slot values with manual efforts.
\end{itemize}

\paragraph{Annotation Task}
\begin{itemize}
  \item \textbf{UtteranceAcceptable}: Is the utterance translation acceptable? Annotate as 1 if the translated utterance is acceptable, 0 otherwise.
  \item \textbf{SlotValuesMatchAcceptable}: Do the translated slot values match the translated utterance? Annotate as 1 if the translated slot values are acceptable, 0 otherwise.
  \item \textbf{NOTE}: Extra notes of judgement.
\end{itemize}

\section{Annotation Example}
Small grammatical errors, but still catch the meaning will be considered \textit{acceptable}. However, if the whole meaning regarding the translation change, it will then be considered as \textit{not acceptable}.

\paragraph{Example 1: Ambiguity}\mbox{} \\
The following example shows the ambiguity issues regarding the translated utterance. In German, \textit{table} can be translated into \textit{Tabelle} as a table form or \textit{Tisch} as a table for reservation. Regarding the contextual information from the utterance, the correct translation should be \textit{Tisch} instead of \textit{Tabelle} in this case. Therefore, the translated utterance will be considered as not acceptable, and annotated as 0. \vspace{0.7em}

\noindent\fbox{%
    \parbox{0.47\textwidth}{%
    \vspace{0.3em}
        \textbf{dialogID}: \textit{PMUL2464.json}\\
        \textbf{turnID}: \textit{9} \\
        \textbf{utterance}: \textit{Yes, Bedouin is a restaurant that serves African food in the Centre.  It is in the expensive range.  Would you like to book a \underline{table}?}\\
        \textbf{slotValues}: {\{restaurant-name: \textit{bedouin}\}}\\
        \textbf{fixTransUtterance}: \textit{Ja, Beduine ist ein Restaurant, das afrikanisches Essen im Zentrum serviert. Es liegt im teuren Bereich. Möchten Sie eine \underline{Tabelle} reservieren?}\\
        \textbf{fixTransSlotValues}: {\{restaurant-name: \textit{Beduine}\}}
        \vspace{0.7em}
        \hrule
        \vspace{0.7em}
        \textbf{UtteranceAcceptable}: 0\\
        \textbf{SlotValuesMatchAcceptable}: 1\\
    }%
}

\paragraph{Example 2: Grammatical Error}\mbox{} \\
The following example shows a slight grammatical issue regarding the translated utterance. This is mainly with the synonym case in Chinese, where the \textit{place} can be translated into \zh{地方} or \zh{位置}, while \zh{位置} will be more appropriate in this scenario. However, \zh{地方} still keep the semantic meaning.  Therefore, the translated utterance will be considered as acceptable, and annotated as 1. And further checking with the translated slot values, all are correct, and should be annotated as 1.\vspace{0.7em}

\noindent\fbox{%
    \parbox{0.47\textwidth}{%
    \vspace{0.3em}
        \textbf{dialogID}: \textit{PMUL0400.json}\\
        \textbf{turnID}: \textit{12} \\
        \textbf{utterance}: \textit{Please book the \underline{place} for 7 people at 11:30 on the same day.}\\
        \textbf{slotValues}: {\{restaurant-people: \textit{7}, restaurant-time: \textit{11:30}, restaurant-day: \textit{Monday}\}}\\
        \textbf{fixTransUtterance}: \zh{请于当天11:30预订7人的\underline{地方}。}\\
        \textbf{fixTransSlotValues}: {\{restaurant-people: \textit{7}, restaurant-time: \textit{11:30}, restaurant-day: \zh{周一}\}}
        \vspace{0.7em}
        \hrule
        \vspace{0.7em}
        \textbf{UtteranceAcceptable}: 1\\
        \textbf{SlotValuesMatchAcceptable}: 1\\
    }%
}

\section{Additional Notes}
Please ignore the slot values with ``dontcare'', ``not mentioned'' and ``none'', while checking the translation quality. If there are any open questions that you think are not covered in this guide, please do not hesitate to get in touch with me or post the questions on Slack, so these issues can be discussed together with other annotators and the guide can be improved.

\clearpage
\onecolumn
\setcounter{section}{2}
\renewcommand{\thesection}{\Alph{section}}
\section{Additional Experiments}

\begin{table*}[htp!]
\centering
{\small
\begin{tabular}{clcccccccccc}
\toprule
 &  & \multicolumn{5}{c}{\textbf{DST}} & \multicolumn{5}{c}{\textbf{RR}} \\
\cmidrule(lr){3-7} \cmidrule(lr){8-12}
\textbf{Lang} & \textbf{Model} & \textbf{1\%} & \textbf{5\%} & \textbf{10\%} & \textbf{50\%} & \textbf{100\%} & \textbf{1\%} & \textbf{5\%} & \textbf{10\%} & \textbf{50\%} & \textbf{100\%} \\
\midrule
\multirow{4}{*}{\textbf{DE}} & TOD-XLMR & 7.68 & 19.26 & 28.08 & 33.17 & 34.10 & 10.25 & 32.47 & 35.56 & 45.39 & 49.46 \\
 & MLM on Mono-CC & 13.75 & 25.15 & 34.12 & 38.01 & 38.26 & 34.37 & 42.13 & 43.51 & 49.10 & 52.80 \\
 & TLM \hspace{0.08em} on OS & 14.17 & 19.45 & 21.62 & 27.28 & 29.91 & 47.21 & 48.59 & 48.96 & 53.01 & 55.30 \\
 & TLM+RS-Mono on OS & 15.88 & 24.14 & 28.38 & 32.57 & 35.45 & 46.08 & 48.94 & 49.98 & 53.43 & 55.72 \\
  \midrule
\multirow{4}{*}{\textbf{AR}} & TOD-XLMR & 1.48 & 1.57 & 6.18 & 15.62 & 17.63 & 6.36 & 18.72 & 23.57 & 36.04 & 42.69 \\
 & MLM on Mono-CC & 4.41 & 5.74 & 7.02 & 14.10 & 17.22 & 28.54 & 31.50 & 32.82 & 41.09 & 44.26 \\
 & TLM \hspace{0.08em} on OS & 4.18 & 6.33 & 6.89 & 13.60 & 17.77 & 32.19 & 35.04 & 37.02 & 41.39 & 47.04 \\
 & TLM+RS-Mono on OS & 4.42 & 6.79 & 8.27 & 14.39 & 21.48 & 33.45 & 37.09 & 38.01 & 41.89 & 47.15 \\
  \midrule
\multirow{4}{*}{\textbf{ZH}} & TOD-XLMR & 8.63 & 12.55 & 16.40 & 23.45 & 25.49 & 15.69 & 31.10 & 33.22 & 41.97 & 48.14 \\
 & MLM on Mono-CC & 11.64 & 19.73 & 25.46 & 34.93 & 35.61 & 34.40 & 37.65 & 39.65 & 48.01 & 50.97 \\
 & TLM \hspace{0.08em} on OS & 11.48 & 17.43 & 21.95 & 28.52 & 32.51 & 38.17 & 42.82 & 42.91 & 49.29 & 51.63 \\
 & TLM+RS-Mono on OS & 11.63 & 14.90 & 17.97 & 22.81 & 28.84 & 38.45 & 43.71 & 45.27 & 48.50 & 51.81 \\
  \midrule
\multirow{4}{*}{\textbf{RU}} & TOD-XLMR & 4.34 & 21.89 & 30.01 & 37.58 & 37.61 & 8.90 & 31.31 & 34.51 & 43.33 & 47.45 \\
 & MLM on Mono-CC & 12.70 & 16.56 & 19.45 & 24.58 & 25.90 & 37.43 & 42.80 & 46.19 & 52.43 & 53.73 \\
 & TLM \hspace{0.08em} on OS & 12.45 & 14.26 & 16.10 & 21.13 & 27.04 & 42.23 & 44.40 & 44.78 & 49.43 & 53.76 \\
 & TLM+RS-Mono on OS & 13.74 & 17.44 & 18.63 & 24.33 & 29.15 & 41.97 & 45.44 & 46.02 & 49.90 & 53.16\\
 \bottomrule
\end{tabular}%
}
\caption{Full per-language few-shot cross-lingual transfer results for Dialog State Tracking and Response Retrieval. Results shown for different sizes of the training data in the target-language (i.e., different number of \textit{shots}): 1\%, 5\%, 10\%, 50\% and 100\% of the \mzs development sets (of respective target languages).}
\label{tab:dst_rr_few_shot_ratio}
\end{table*}

\end{document}